\documentclass[journal]{IEEEtran}
\IEEEoverridecommandlockouts
\usepackage{cite}
\usepackage{amsmath,amssymb,amsfonts}
\usepackage{algorithmic}
\usepackage{graphicx}
\usepackage{textcomp}
\usepackage{xcolor}
\def\BibTeX{{\rm B\kern-.05em{\sc i\kern-.025em b}\kern-.08em
    T\kern-.1667em\lower.7ex\hbox{E}\kern-.125emX}}

\def\PsfigVersion{1.9}
\ifx\undefined\psfig\else \fi

\let\LaTeXAtSign=\@
\let\@=\relax
\edef\psfigRestoreAt{\catcode`\@=\number\catcode`@\relax}
\catcode`\@=11\relax
\newwrite\@unused
\def\ps@typeout#1{{\let\protect\string\immediate\write\@unused{#1}}}
\ps@typeout{psfig/tex \PsfigVersion}

\def\figurepath{./}

\def\@nnil{\@nil}
\def\@empty{}
\def\@psdonoop#1\@@#2#3{}
\def\@psdo#1:=#2\do#3{\edef\@psdotmp{#2}\ifx\@psdotmp\@empty \else
    \expandafter\@psdoloop#2,\@nil,\@nil\@@#1{#3}\fi}
\def\@psdoloop#1,#2,#3\@@#4#5{\def#4{#1}\ifx #4\@nnil \else
       #5\def#4{#2}\ifx #4\@nnil \else#5\@ipsdoloop #3\@@#4{#5}\fi\fi}
\def\@ipsdoloop#1,#2\@@#3#4{\def#3{#1}\ifx #3\@nnil 
       \let\@nextwhile=\@psdonoop \else
      #4\relax\let\@nextwhile=\@ipsdoloop\fi\@nextwhile#2\@@#3{#4}}
\def\@tpsdo#1:=#2\do#3{\xdef\@psdotmp{#2}\ifx\@psdotmp\@empty \else
    \@tpsdoloop#2\@nil\@nil\@@#1{#3}\fi}
\def\@tpsdoloop#1#2\@@#3#4{\def#3{#1}\ifx #3\@nnil 
       \let\@nextwhile=\@psdonoop \else
      #4\relax\let\@nextwhile=\@tpsdoloop\fi\@nextwhile#2\@@#3{#4}}
\ifx\undefined\fbox
\newdimen\fboxrule
\newdimen\fboxsep
\newdimen\ps@tempdima
\newbox\ps@tempboxa
\long\def\fbox#1{\leavevmode\setbox\ps@tempboxa\hbox{#1}\ps@tempdima\fboxrule
    \advance\ps@tempdima \fboxsep \advance\ps@tempdima \dp\ps@tempboxa
   \hbox{\lower \ps@tempdima\hbox
  {\vbox{\hrule height \fboxrule
          \hbox{\vrule width \fboxrule \hskip\fboxsep
          \vbox{\vskip\fboxsep \box\ps@tempboxa\vskip\fboxsep}\hskip 
                 \fboxsep\vrule width \fboxrule}
                 \hrule height \fboxrule}}}}
\fi
\newread\ps@stream
\newif\ifnot@eof       
\newif\if@noisy        
\newif\if@atend        
\newif\if@psfile       
{\catcode`\%=12\global\gdef\epsf@start{
\def\epsf@PS{PS}
\def\epsf@getbb#1{%
\openin\ps@stream=#1
\ifeof\ps@stream\ps@typeout{Error, File #1 not found}\else
   {\not@eoftrue \chardef\other=12
    \def\do##1{\catcode`##1=\other}\dospecials \catcode`\ =10
    \loop
       \if@psfile
	  \read\ps@stream to \epsf@fileline
       \else{
	  \obeyspaces
          \read\ps@stream to \epsf@tmp\global\let\epsf@fileline\epsf@tmp}
       \fi
       \ifeof\ps@stream\not@eoffalse\else
       \if@psfile\else
       \expandafter\epsf@test\epsf@fileline:. \\%
       \fi
          \expandafter\epsf@aux\epsf@fileline:. \\%
       \fi
   \ifnot@eof\repeat
   }\closein\ps@stream\fi}%
\long\def\epsf@test#1#2#3:#4\\{\def\epsf@testit{#1#2}
			\ifx\epsf@testit\epsf@start\else
\ps@typeout{Warning! File does not start with `\epsf@start'.  It may not be a PostScript file.}
			\fi
			\@psfiletrue} 
{\catcode`\%=12\global\let\epsf@percent=
\long\def\epsf@aux#1#2:#3\\{\ifx#1\epsf@percent
   \def\epsf@testit{#2}\ifx\epsf@testit\epsf@bblit
	\@atendfalse
        \epsf@atend #3 . \\%
	\if@atend	
	   \if@verbose{
		\ps@typeout{psfig: found `(atend)'; continuing search}
	   }\fi
        \else
        \epsf@grab #3 . . . \\%
        \not@eoffalse
        \global\no@bbfalse
        \fi
   \fi\fi}%
\def\epsf@grab #1 #2 #3 #4 #5\\{%
   \global\def\epsf@llx{#1}\ifx\epsf@llx\empty
      \epsf@grab #2 #3 #4 #5 .\\\else
   \global\def\epsf@lly{#2}%
   \global\def\epsf@urx{#3}\global\def\epsf@ury{#4}\fi}%
\def\epsf@atendlit{(atend)} 
\def\epsf@atend #1 #2 #3\\{%
   \def\epsf@tmp{#1}\ifx\epsf@tmp\empty
      \epsf@atend #2 #3 .\\\else
   \ifx\epsf@tmp\epsf@atendlit\@atendtrue\fi\fi}

\chardef\psletter = 11 
\chardef\other = 12

\newif \ifdebug 
\newif\ifc@mpute 
\c@mputetrue 

\let\then = \relax
\def\r@dian{pt }
\let\r@dians = \r@dian
\let\dimensionless@nit = \r@dian
\let\dimensionless@nits = \dimensionless@nit
\def\internal@nit{sp }
\let\internal@nits = \internal@nit
\newif\ifstillc@nverging
\def \Mess@ge #1{\ifdebug \then \message {#1} \fi}

{ 
	\catcode `\@ = \psletter
	\gdef \nodimen {\expandafter \n@dimen \the \dimen}
	\gdef \term #1 #2 #3%
	       {\edef \t@ {\the #1}
		\edef \t@@ {\expandafter \n@dimen \the #2\r@dian}%
		\t@rm {\t@} {\t@@} {#3}%
	       }
	\gdef \t@rm #1 #2 #3%
	       {{%
		\count 0 = 0
		\dimen 0 = 1 \dimensionless@nit
		\dimen 2 = #2\relax
		\Mess@ge {Calculating term #1 of \nodimen 2}%
		\loop
		\ifnum	\count 0 < #1
		\then	\advance \count 0 by 1
			\Mess@ge {Iteration \the \count 0 \space}%
			\Multiply \dimen 0 by {\dimen 2}%
			\Mess@ge {After multiplication, term = \nodimen 0}%
			\Divide \dimen 0 by {\count 0}%
			\Mess@ge {After division, term = \nodimen 0}%
		\repeat
		\Mess@ge {Final value for term #1 of 
				\nodimen 2 \space is \nodimen 0}%
		\xdef \Term {#3 = \nodimen 0 \r@dians}%
		\aftergroup \Term
	       }}
	\catcode `\p = \other
	\catcode `\t = \other
	\gdef \n@dimen #1pt{#1} 
}

\def \Divide #1by #2{\divide #1 by #2} 

\def \Multiply #1by #2
       {{
	\count 0 = #1\relax
	\count 2 = #2\relax
	\count 4 = 65536
	\Mess@ge {Before scaling, count 0 = \the \count 0 \space and
			count 2 = \the \count 2}%
	\ifnum	\count 0 > 32767 
	\then	\divide \count 0 by 4
		\divide \count 4 by 4
	\else	\ifnum	\count 0 < -32767
		\then	\divide \count 0 by 4
			\divide \count 4 by 4
		\else
		\fi
	\fi
	\ifnum	\count 2 > 32767 
	\then	\divide \count 2 by 4
		\divide \count 4 by 4
	\else	\ifnum	\count 2 < -32767
		\then	\divide \count 2 by 4
			\divide \count 4 by 4
		\else
		\fi
	\fi
	\multiply \count 0 by \count 2
	\divide \count 0 by \count 4
	\xdef \product {#1 = \the \count 0 \internal@nits}%
	\aftergroup \product
       }}

\def\r@duce{\ifdim\dimen0 > 90\r@dian \then   
		\multiply\dimen0 by -1
		\advance\dimen0 by 180\r@dian
		\r@duce
	    \else \ifdim\dimen0 < -90\r@dian \then  
		\advance\dimen0 by 360\r@dian
		\r@duce
		\fi
	    \fi}

\def\Sine#1%
       {{%
	\dimen 0 = #1 \r@dian
	\r@duce
	\ifdim\dimen0 = -90\r@dian \then
	   \dimen4 = -1\r@dian
	   \c@mputefalse
	\fi
	\ifdim\dimen0 = 90\r@dian \then
	   \dimen4 = 1\r@dian
	   \c@mputefalse
	\fi
	\ifdim\dimen0 = 0\r@dian \then
	   \dimen4 = 0\r@dian
	   \c@mputefalse
	\fi
	\ifc@mpute \then
		\divide\dimen0 by 180
		\dimen0=3.141592654\dimen0
		\dimen 2 = 3.1415926535897963\r@dian 
		\divide\dimen 2 by 2 
		\Mess@ge {Sin: calculating Sin of \nodimen 0}%
		\count 0 = 1 
		\dimen 2 = 1 \r@dian 
		\dimen 4 = 0 \r@dian 
		\loop
			\ifnum	\dimen 2 = 0 
			\then	\stillc@nvergingfalse 
			\else	\stillc@nvergingtrue
			\fi
			\ifstillc@nverging 
			\then	\term {\count 0} {\dimen 0} {\dimen 2}%
				\advance \count 0 by 2
				\count 2 = \count 0
				\divide \count 2 by 2
				\ifodd	\count 2 
				\then	\advance \dimen 4 by \dimen 2
				\else	\advance \dimen 4 by -\dimen 2
				\fi
		\repeat
	\fi		
			\xdef \sine {\nodimen 4}%
       }}

\def\Cosine#1{\ifx\sine\UnDefined\edef\Savesine{\relax}\else
		             \edef\Savesine{\sine}\fi
	{\dimen0=#1\r@dian\advance\dimen0 by 90\r@dian
	 \Sine{\nodimen 0}
	 \xdef\cosine{\sine}
	 \xdef\sine{\Savesine}}}	      

\def\psdraft{
	\def\@psdraft{0}
}
\def\psfull{
	\def\@psdraft{100}
}

\psfull

\newif\if@scalefirst
\def\psscalefirst{\@scalefirsttrue}
\def\psrotatefirst{\@scalefirstfalse}
\psrotatefirst

\newif\if@draftbox
\def\psnodraftbox{
	\@draftboxfalse
}
\def\psdraftbox{
	\@draftboxtrue
}
\@draftboxtrue

\newif\if@prologfile
\newif\if@postlogfile
\def\pssilent{
	\@noisyfalse
}
\def\psnoisy{
	\@noisytrue
}
\psnoisy
\newif\if@bbllx
\newif\if@bblly
\newif\if@bburx
\newif\if@bbury
\newif\if@height
\newif\if@width
\newif\if@rheight
\newif\if@rwidth
\newif\if@angle
\newif\if@clip
\newif\if@verbose
\def\@p@@sclip#1{\@cliptrue}

\newif\if@decmpr

\def\@p@@sfigure#1{\def\@p@sfile{null}\def\@p@sbbfile{null}
	        \openin1=#1.bb
		\ifeof1\closein1
	        	\openin1=\figurepath#1.bb
			\ifeof1\closein1
			        \openin1=#1
				\ifeof1\closein1%
				       \openin1=\figurepath#1
					\ifeof1
					   \ps@typeout{Error, File #1 not found}
						\if@bbllx\if@bblly
				   		\if@bburx\if@bbury
			      				\def\@p@sfile{#1}%
			      				\def\@p@sbbfile{#1}%
							\@decmprfalse
				  	   	\fi\fi\fi\fi
					\else\closein1
				    		\def\@p@sfile{\figurepath#1}%
				    		\def\@p@sbbfile{\figurepath#1}%
						\@decmprfalse
	                       		\fi%
			 	\else\closein1%
					\def\@p@sfile{#1}
					\def\@p@sbbfile{#1}
					\@decmprfalse
			 	\fi
			\else
				\def\@p@sfile{\figurepath#1}
				\def\@p@sbbfile{\figurepath#1.bb}
				\@decmprtrue
			\fi
		\else
			\def\@p@sfile{#1}
			\def\@p@sbbfile{#1.bb}
			\@decmprtrue
		\fi}

\def\@p@@sfile#1{\@p@@sfigure{#1}}

\def\@p@@sbbllx#1{
		\@bbllxtrue
		\dimen100=#1
		\edef\@p@sbbllx{\number\dimen100}
}
\def\@p@@sbblly#1{
		\@bbllytrue
		\dimen100=#1
		\edef\@p@sbblly{\number\dimen100}
}
\def\@p@@sbburx#1{
		\@bburxtrue
		\dimen100=#1
		\edef\@p@sbburx{\number\dimen100}
}
\def\@p@@sbbury#1{
		\@bburytrue
		\dimen100=#1
		\edef\@p@sbbury{\number\dimen100}
}
\def\@p@@sheight#1{
		\@heighttrue
		\dimen100=#1
   		\edef\@p@sheight{\number\dimen100}
}
\def\@p@@swidth#1{
		\@widthtrue
		\dimen100=#1
		\edef\@p@swidth{\number\dimen100}
}
\def\@p@@srheight#1{
		\@rheighttrue
		\dimen100=#1
		\edef\@p@srheight{\number\dimen100}
}
\def\@p@@srwidth#1{
		\@rwidthtrue
		\dimen100=#1
		\edef\@p@srwidth{\number\dimen100}
}
\def\@p@@sangle#1{
		\@angletrue
		\edef\@p@sangle{#1} 
}
\def\@p@@ssilent#1{ 
		\@verbosefalse
}
\def\@p@@sprolog#1{\@prologfiletrue\def\@prologfileval{#1}}
\def\@p@@spostlog#1{\@postlogfiletrue\def\@postlogfileval{#1}}
\def\@cs@name#1{\csname #1\endcsname}
\def\@setparms#1=#2,{\@cs@name{@p@@s#1}{#2}}
\def\ps@init@parms{
		\@bbllxfalse \@bbllyfalse
		\@bburxfalse \@bburyfalse
		\@heightfalse \@widthfalse
		\@rheightfalse \@rwidthfalse
		\def\@p@sbbllx{}\def\@p@sbblly{}
		\def\@p@sbburx{}\def\@p@sbbury{}
		\def\@p@sheight{}\def\@p@swidth{}
		\def\@p@srheight{}\def\@p@srwidth{}
		\def\@p@sangle{0}
		\def\@p@sfile{} \def\@p@sbbfile{}
		\def\@p@scost{10}
		\def\@sc{}
		\@prologfilefalse
		\@postlogfilefalse
		\@clipfalse
		\if@noisy
			\@verbosetrue
		\else
			\@verbosefalse
		\fi
}
\def\parse@ps@parms#1{
	 	\@psdo\@psfiga:=#1\do
		   {\expandafter\@setparms\@psfiga,}}
\newif\ifno@bb
\def\bb@missing{
	\if@verbose{
		\ps@typeout{psfig: searching \@p@sbbfile \space  for bounding box}
	}\fi
	\no@bbtrue
	\epsf@getbb{\@p@sbbfile}
        \ifno@bb \else \bb@cull\epsf@llx\epsf@lly\epsf@urx\epsf@ury\fi
}	
\def\bb@cull#1#2#3#4{
	\dimen100=#1 bp\edef\@p@sbbllx{\number\dimen100}
	\dimen100=#2 bp\edef\@p@sbblly{\number\dimen100}
	\dimen100=#3 bp\edef\@p@sbburx{\number\dimen100}
	\dimen100=#4 bp\edef\@p@sbbury{\number\dimen100}
	\no@bbfalse
}
\newdimen\p@intvaluex
\newdimen\p@intvaluey
\def\rotate@#1#2{{\dimen0=#1 sp\dimen1=#2 sp
		  \global\p@intvaluex=\cosine\dimen0
		  \dimen3=\sine\dimen1
		  \global\advance\p@intvaluex by -\dimen3
		  \global\p@intvaluey=\sine\dimen0
		  \dimen3=\cosine\dimen1
		  \global\advance\p@intvaluey by \dimen3
		  }}
\def\compute@bb{
		\no@bbfalse
		\if@bbllx \else \no@bbtrue \fi
		\if@bblly \else \no@bbtrue \fi
		\if@bburx \else \no@bbtrue \fi
		\if@bbury \else \no@bbtrue \fi
		\ifno@bb \bb@missing \fi
		\ifno@bb \ps@typeout{FATAL ERROR: no bb supplied or found}
			\no-bb-error
		\fi
		\count203=\@p@sbburx
		\count204=\@p@sbbury
		\advance\count203 by -\@p@sbbllx
		\advance\count204 by -\@p@sbblly
		\edef\ps@bbw{\number\count203}
		\edef\ps@bbh{\number\count204}
		\if@angle 
			\Sine{\@p@sangle}\Cosine{\@p@sangle}
	        	{\dimen100=\maxdimen\xdef\r@p@sbbllx{\number\dimen100}
					    \xdef\r@p@sbblly{\number\dimen100}
			                    \xdef\r@p@sbburx{-\number\dimen100}
					    \xdef\r@p@sbbury{-\number\dimen100}}
                        \def\minmaxtest{
			   \ifnum\number\p@intvaluex<\r@p@sbbllx
			      \xdef\r@p@sbbllx{\number\p@intvaluex}\fi
			   \ifnum\number\p@intvaluex>\r@p@sbburx
			      \xdef\r@p@sbburx{\number\p@intvaluex}\fi
			   \ifnum\number\p@intvaluey<\r@p@sbblly
			      \xdef\r@p@sbblly{\number\p@intvaluey}\fi
			   \ifnum\number\p@intvaluey>\r@p@sbbury
			      \xdef\r@p@sbbury{\number\p@intvaluey}\fi
			   }
			\rotate@{\@p@sbbllx}{\@p@sbblly}
			\minmaxtest
			\rotate@{\@p@sbbllx}{\@p@sbbury}
			\minmaxtest
			\rotate@{\@p@sbburx}{\@p@sbblly}
			\minmaxtest
			\rotate@{\@p@sbburx}{\@p@sbbury}
			\minmaxtest
			\edef\@p@sbbllx{\r@p@sbbllx}\edef\@p@sbblly{\r@p@sbblly}
			\edef\@p@sbburx{\r@p@sbburx}\edef\@p@sbbury{\r@p@sbbury}
		\fi
		\count203=\@p@sbburx
		\count204=\@p@sbbury
		\advance\count203 by -\@p@sbbllx
		\advance\count204 by -\@p@sbblly
		\edef\@bbw{\number\count203}
		\edef\@bbh{\number\count204}
}
\def\in@hundreds#1#2#3{\count240=#2 \count241=#3
		     \count100=\count240	
		     \divide\count100 by \count241
		     \count101=\count100
		     \multiply\count101 by \count241
		     \advance\count240 by -\count101
		     \multiply\count240 by 10
		     \count101=\count240	
		     \divide\count101 by \count241
		     \count102=\count101
		     \multiply\count102 by \count241
		     \advance\count240 by -\count102
		     \multiply\count240 by 10
		     \count102=\count240	
		     \divide\count102 by \count241
		     \count200=#1\count205=0
		     \count201=\count200
			\multiply\count201 by \count100
		 	\advance\count205 by \count201
		     \count201=\count200
			\divide\count201 by 10
			\multiply\count201 by \count101
			\advance\count205 by \count201
		     \count201=\count200
			\divide\count201 by 100
			\multiply\count201 by \count102
			\advance\count205 by \count201
		     \edef\@result{\number\count205}
}
\def\compute@wfromh{
		\in@hundreds{\@p@sheight}{\@bbw}{\@bbh}
		\edef\@p@swidth{\@result}
}
\def\compute@hfromw{
	        \in@hundreds{\@p@swidth}{\@bbh}{\@bbw}
		\edef\@p@sheight{\@result}
}
\def\compute@handw{
		\if@height 
			\if@width
			\else
				\compute@wfromh
			\fi
		\else 
			\if@width
				\compute@hfromw
			\else
				\edef\@p@sheight{\@bbh}
				\edef\@p@swidth{\@bbw}
			\fi
		\fi
}
\def\compute@resv{
		\if@rheight \else \edef\@p@srheight{\@p@sheight} \fi
		\if@rwidth \else \edef\@p@srwidth{\@p@swidth} \fi
}
\def\compute@sizes{
	\compute@bb
	\if@scalefirst\if@angle
	\if@width
	   \in@hundreds{\@p@swidth}{\@bbw}{\ps@bbw}
	   \edef\@p@swidth{\@result}
	\fi
	\if@height
	   \in@hundreds{\@p@sheight}{\@bbh}{\ps@bbh}
	   \edef\@p@sheight{\@result}
	\fi
	\fi\fi
	\compute@handw
	\compute@resv}

\def\psfig#1{\vbox {
	\ps@init@parms
	\parse@ps@parms{#1}
	\compute@sizes
	\ifnum\@p@scost<\@psdraft{
		\special{ps::[begin] 	\@p@swidth \space \@p@sheight \space
				\@p@sbbllx \space \@p@sbblly \space
				\@p@sbburx \space \@p@sbbury \space
				startTexFig \space }
		\if@angle
			\special {ps:: \@p@sangle \space rotate \space} 
		\fi
		\if@clip{
			\if@verbose{
				\ps@typeout{(clip)}
			}\fi
			\special{ps:: doclip \space }
		}\fi
		\if@prologfile
		    \special{ps: plotfile \@prologfileval \space } \fi
		\if@decmpr{
			\if@verbose{
				\ps@typeout{psfig: including \@p@sfile.Z \space }
			}\fi
			\special{ps: plotfile "`zcat \@p@sfile.Z" \space }
		}\else{
			\if@verbose{
				\ps@typeout{psfig: including \@p@sfile \space }
			}\fi
			\special{ps: plotfile \@p@sfile \space }
		}\fi
		\if@postlogfile
		    \special{ps: plotfile \@postlogfileval \space } \fi
		\special{ps::[end] endTexFig \space }
		\vbox to \@p@srheight sp{
			\hbox to \@p@srwidth sp{
				\hss
			}
		\vss
		}
	}\else{
		\if@draftbox{		
			\hbox{\frame{\vbox to \@p@srheight sp{
			\vss
			\hbox to \@p@srwidth sp{ \hss \@p@sfile \hss }
			\vss
			}}}
		}\else{
			\vbox to \@p@srheight sp{
			\vss
			\hbox to \@p@srwidth sp{\hss}
			\vss
			}
		}\fi

	}\fi
}}
\psfigRestoreAt
\let\@=\LaTeXAtSign

\begin{document}

\title{A Large Language Model-based Computational Approach to Improve Identity-Related Write-Ups}

\author{Alex Doboli \\  Department of Electrical and Computer Engineering \\ Stony Brook University, Stony Brook, NY \\ alex.doboli@stonybrook.edu}
\maketitle

\begin{abstract}
Creating written products is essential to modern life, including writings about one's identity and personal experiences.  However, writing is often a difficult activity that requires extensive effort to frame the central ideas, the pursued approach to communicate the central ideas, e.g., using analogies, metaphors, or other possible means, the needed presentation structure, and the actual verbal expression. Large Language Models, a recently emerged approach in Machine Learning, can offer a significant help in reducing the effort and improving the quality of written products. This paper proposes a new computational approach to explore prompts that given as inputs to a Large Language Models can generate cues to improve the considered written products. Two case studies on improving write-ups, one based on an analogy and one on a metaphor, are also presented in the paper.  
\end{abstract}

\section {Introduction}

Document preparation is one of the most important activities in human life, as it impacts every facet, like solution documenting, application description, situation narration, prediction formulation, opinion statement, social networking, and many more. Moreover, writing about one's identity is critical too, as it expresses one's personal characteristics, including upbringing, experiences, preferences and priorities, and goals and objectives~\cite{Scenters2014}. Hence, the writing process should not only articulate well the main message, e.g., its central idea, involving any identity-related issues that the author intends to communicate, but also do it persuasively.

The variables that determine the quality of persuasive writing include ``coherence, cohesion, syntactic features, and persuasive appeals''~\cite{Connor1985}, pp. 314. For example, Connor and Lauer mention a system of twenty-three appeals grouped into three categories: rational appeals, credibility appeals, and affective appeals~\cite{Connor1985}. The appeals included in the three categories are as follows: ``rational appeals are descriptive examples, narrative examples, classifications, comparisons, contrasts, degree, and authority; credibility appeals are firsthand experience, writer's respect for audience's interest and point of view, writer-audience shared interests and points of view, and writer's good character and or judgement; and affective appeals are emotion in audience's situation, audience's empathy, audience's values, vivid picture, charged language, cause/effect, model, stage in process, means/end, consequences, ideal or principle, and information (facts, statistics)''~\cite{Connor1985}, pp. 314.   

Persuasively writing to reflect one's identity is difficult. From a cognitive point of view, it requires framing the intended messages while combining its idea with the author's own experiences, situations and perspectives as well as extensive self-analysis and self-inspection. The goal is to identify the central idea to be communicated and the vehicle for its communication, like the narrative solutions. For example, relying on popular analogies and metaphors, like characters and stories in Greek mythology, has been a popular approach to tackle this difficulty, as they arguably offer a frame already known and accepted by the targeted audience. However, the writing activity remains challenging even in such cases, as the author still must figure out the best way of combining the central idea of the intended writing with such vehicles, the structuring of the writing, and its actual presentation. It is often the case that multiple rewriting as required, often with substantial modifications, in order to optimize the three enumerated elements. Devising new ways, possibly supported by computational tools, are needed not only to effectively manage the writing process, like the required time and effort, but also to improve the message, its correctness, and expected understanding by the audience.          

Large Language Models (LLMs) have recently emerged as a very promising approach to tackle situations expressed in natural language (NL), in a broad variety of domains including education, programming, medicine, and so on~\cite{Thirunavukarasu2023}. LLMs arguably describe the solution space formed by the ideas expressed by humans in writing. LLMs are broad collections of concepts (e.g., words) linked together based on their co-occurrences in huge sets of electronic documents. Exploring this solution space has been achieved through sequences of prompts, which express certain requirements in NL. LLM inquiries through prompts produce responses, also in NL. However, generating effective sequences of prompts to address an author's intentions about the messages to be communicated remains difficult. While the area of prompt engineering has started to address the related challenges, the responses produced by LLMs tend to be produced in a certain structure, like using enumerations, mainly focus on a few well-known hence possibly repetitive examples, which can reduce the creativity of a writing, and have limited ability in combining separate ideas (e.g., beyond only mixing separate sentences about the combined ideas), as needed when attempting to intertwine one's identity-related aspects with the narrative story.   

This paper presents a computational approach to improve the message and presentation quality of written products, e.g., write-ups, based on the cues produced by Large Language Models in response to user generated prompts. The proposed approach is based on the observation that cue exploration includes the following activities: (a)~finding the concepts and ideas related to a write-up; (b)~detailing the concepts and ideas of the write-up; (c)~evaluating the available cues; (d)~selecting the cues to be further explored among the available options, i.e. related and detailed concepts and ideas, (e)~combining the concepts and ideas of the write-up and selected cues, and (f)~pursuing parallel exploration threads for different ideas. The proposed computational approach implements activities~(a)-(f) through an iterative exploration loop that groups the available cues into cues to be explored, cues used to evaluate the current write-up, and cues to be ignored; prioritizing the selected cues using the cues for evaluation; separately exploring the selected cues based on their priority; combining the cues of separate (parallel) explorations with those of detailing; and generating new prompts for further detailing of the ideas. Activities~(a)-(f) and the related computational approach to improve written products were devised based on the insight gained from processing two separate write-ups to reflect the author's identity, i.e. passion for dancing, one based on the analogy between dancing and poetry, and the second using the metaphor of the story of Icarus. Both examples are detailed in the paper.

The paper has the following structure. Related work is discussed in Section~II. Section~III discusses an example based on analogy
making. Section~IV presents an example using a metaphor. Section~V describes the proposed algorithmic approach to improve writing using LLMs, and Section~VI proposes a set of related evaluation metrics. Section~VII details related experiments on using the proposed algorithmic approach for the two examples. Conclusions end the paper.  

\section {Related Work}

The existing approaches that use LLMs to solve a need include the following activities: (i)~prompt engineering to create new prompts, (ii)~LLM exploration through sequences of prompts, and visualization of the explored space to understand the responses obtained for the offered prompts. LLM exploration is significantly different from traditional, numerical exploration, as LLMs are ultimately grounded in the semantics of NL, while traditional optimization is based on theories in mathematical calculus. The above three activities are discussed next.    

Prompt engineering is the activity of devising the required prompts, so that the desired outputs are being obtained~\cite{Spinner2023, Webson2022}. The main challenge pertains to understanding the obtained results, which are expressed in NL, so that new cues can automatically be generated. Hence, traditional challenges in understanding the semantics of NL, including understanding ambiguities, uncertainties, analogies, metaphors, and so on, are important in prompt engineering too. Interpretability and explainability of LLM has been recently studied~\cite{Strobelt2019, Strobelt2022}. Related difficulties include LLM biasing and prompt sensitivity~\cite{Spinner2023}. Spinner et al. classify the challenges of LLM into three kinds, (i)~data- and model-specific, (ii)~linguistic, and (iii)~socio-linguistic challenges~\cite{Spinner2023}. The enumerated challenges include sensitivity to small changes in prompt formulations, surface form competition that lowers the probability distribution for semantically-similar words, handling of negation and quantifiers, and biasing~\cite{Spinner2023}. Tackling prompt sensitivity requires understanding the causality relationships on how prompt changes produce output variations~\cite{Almeda2023}. A consequence is that the description of the utilized prompts must be varied and multiple response candidates should be considered.  

Prompt generation is the process of creating a new prompt by combining a prompt structure with a word set, e.g., word embeddings. Both prompt structures and word sets are selected using an input, which can be another text message or an image. A prompt structure, often called prompt in the related literature, is a parameterized template, in which the parameters are linked to the words of the selected embeddings. The existing prompt generation methods differ depending on how prompt structures and word embeddings are learned and selected. A category of methods use Bayesian methods to create new prompts~\cite{Derakhshani2023}. Contrastive Language-Image Pretraining (CLIP) uses manually crafted prompt structures selected using a Bayesian classifier~\cite{Radford2021}. The limitation of having to hand craft prompt structures is addressed in Context optimization, which learns embedded matrices for prompt structures obtained using CLIP method~\cite{Zhou2022}. Conditional Prompt Learning trains a neural network to output prompt structures for an input image~\cite{Zhou2022b}. Similarly, a Gaussian distribution conditioned by an input image is used to sample a set of residuals that are then combined with class-specific embeddings to find the most probable prompt~\cite{Derakhshani2023}. The method in~\cite{Zhang2022} uses an encoder and autoregression decoder based on a transformer architecture. The encoder represents noun associations for different objectives described as masks to create a hidden state. The decoder uses the hidden state to generate a new prompt. Its quality is evaluated using negative log-like function. 

Design space exploration for text-to-image exploration has been discussed in~\cite{Almeda2023}. Prompt sensitivity, like image changes due to small prompt variations, has been enumerated among the current challenges. Exploration uses a set of LLM-based functions for generating an image using the given prompt, obtaining an arbitrary prompt, populating the cells in a row (or
with words and phrases of type prompt, producing a list of synonyms, creating a list of antonyms, generating ``divergent'' words, creating a list of alternative wordings for prompt,
and producing an embellished alternative to prompt by using
more specific or detailed words~\cite{Almeda2023}. Exploration should consider multiple alternatives~\cite{Zaman2015}. Chain-of-thought (CoT) prompting first decomposes a problem into separate questions, which are then answered in sequence~\cite{Zhang2022b}. One approach prompts at every step for reasoning steps, e.g., using the prompt ``Let's think step by step ...''~\cite{Wei2023}. A better approach is to use at each step a question, a reasoning chain, and an expected answer. Finally, a solution to automate CoT prompting groups sequences of questions into clusters, and then generates the reasoning chain for the representative question of each cluster~\cite{Zhang2022b}.

Graphologue is an interactive system to translate text-based responses from LLMs into graphical diagrams to aid information search and question answering~\cite{Jiang2023}. It extracts entities and relationships from LLM responses and constructs visual diagrams. The approach called DREAMSHEETS aids in creating the prompts and display of generated results using
spreadsheet interface~\cite{Almeda2023}.

\section {Example 1: Analogy}

The following paragraph is part of the essay ``Dance is Poetry Moving''  by an author for which dancing is a main way of expressing his identity~\cite{Sandy1}. The paragraph suggests that there is an analogy between poetry and dance, and then elaborates on the similarities between the two. GhatGPT was used to improve the understanding and the outcome of the analogy, including its plausibility, and expected emotional impact. 
 
``Some of you might disagree with this and argue that the topic in poetry is analogous to the story in a dance, since both contain the narrative and sentiment of a dance or poem. However, the story in a dance is influenced by the song, while there is no base for a topic in poetry. The song and topic both serve as the base for their respective pieces. The choreography, emotions, and musicality are all based on the type of song implemented. We try to find ways to mimic the lyrics or specific instruments of the music. If there is a bass drum in the background, some of our movements would be definite and tight in order to emphasize it's staccato tone. In poetry, we create verses based on our own central idea. We organize the structure of the poem, such as the number of lines per stanza, based on what we want the audience to remember. We single out specific lines as their own stanzas or in special form, like all capitals or bold, in order to present an additional sentiment to the poem and that line.''

Analogies have been extensively studied in cognitive psychology and in design sciences~\cite{Vosniadou1989}. Understanding analogies implies identifying isomorphic structures among the referred descriptions, including finding the common properties or the similarity degree of the utilized examples in the description~\cite{Ferent2013, Weitzenfeld1984}. Hence, understanding analogies is similar to puzzle solving, in which the complexity of solving depends on uncertainty involving the variables that can be associated and their relations. Weitzenfeld explains that analogies require that the involved variables are connected through similar relations~\cite{Weitzenfeld1984}. Every analogy is characterized by the soundness and plausibility of the argument, e.g., the suggested isomorphism between the variables of the involved ideas, and the utility and expected impact, including the emotional impact, of the isomorphism described by an analogy~\cite{Weitzenfeld1984}. A symbolic matching algorithm to identify analogies in electronic circuits has been suggested in~\cite{Ferent2013}.

There are still no accepted cognitive theories on similarity judgement in analogies~\cite{Weitzenfeld1984}, e.g., how relevance emerges for an idea and how relevance is used in understanding analogies. Also, it is not guaranteed that all readers will ``solve'' in the same way the puzzle of an analogy, which can create a situation in which different readers identify different associations, especially for analogies involving variables from different domains, like poetry and dance in the paragraph above. Contradictions can result during processing of analogies. Using examples to illustrate broad analogies reduces the number of possible associations, and hence increase the robustness of analogy understanding.

The above paragraph describes the proposed analogy, e.g., {\em topic} in poetry and {\em story} in dance, the explanation (support) for the proposed analogy, and any differences that do not fit the analogy. The paragraph then delves into detailing the broad association between poetry and dance into their specifics, and the consequence of these associations. The similarity between topic and story is that both share a narrative to communicate a sentiment. However, there is also a difference between the two: while the story in dance is based on the song, there is no equivalent for a topic in poetry, as the topic serves the role of the base too. Hence, the analogy stops here. The detailing of the analogy starts from the stated insight that both mimic lyrics and instruments of the music. This second, more detailed analogy, which supports the broad analogy between topic and story, serves then as a justification for the detailing of the analogy, e.g., using bass drum in the background for dance, and the structure of verses for poetry. The similarity in structure refers to different variables for the two cases. The analogy includes not only variables and their relationships, but also their outcomes, like the staccato tone in music and the concepts to be remembered by the audience in poetry.

The next subsection discusses the use of ChatGPT to strengthen the desired analogy.

\subsection {Exploration Trace}

The goal of exploration was to identify cues that support the improvement of the above analogy by strengthening (i)~its plausibility, e.g., the soundness of the suggested isomorphism between the two concepts of the analogy, and (ii)~impact, like the expected emotional impact on the reader. Cue exploration includes the following activities: {\tt (a)}~finding related concepts and ideas; {\tt (b)}~detailing existing concepts and ideas; {\tt (c)}~evaluating the available cues; {\tt (d)}~selecting among the available options, i.e. related and detailed concepts and ideas, {\tt (e)}~combining concepts and ideas, and {\tt (f)}~pursuing parallel exploration threads for different ideas (followed by executing item~{\tt (e)} to combine the identified cues).     

First, prompting ChatGPT to comment on the above paragraph (activity~{\tt (c)}) produced a result stating that the quality is fair but that its clarity and coherence could be improved. Specific suggestions included the following seven elements: {\tt (PROMPT1.1)}~clarity of analogies, e.g., their type, explanations, and clarity; {\tt (PROMPT1.2)}~structural flow, i.e. organization, flow, and transition of ideas; {\tt (PROMPT1.3)}~supporting examples, like using concrete examples to support the analogy; {\tt (PROMPT1.4)}~conciseness of the paragraph; {\tt (PROMPT1.5)}~grammar and language; and {\tt (PROMPT1.6)}~formatting the paragraph.

Based on the choice in elaborating the paragraph, the pursued exploration decided to transform the above paragraph by focusing on items~{\tt (PROMPT1.1)} (clarity of analogies) and~{\tt (PROMPT1.3)} (utilizing concrete examples to illustrate the analogies in the paragraph) from the above list created by ChatGPT. The two items were expected to reduce the uncertainties of the analogy, hence strengthen its plausibility. Items~{\tt (PROMPT1.2)} (structural flow) and~{\tt (PROMPT1.4)} (conciseness) were used as metrics to assess the quality of different transformation options, as they are more amenable to quantitative measuring. Assessing the emotional impact of the transformed paragraphs is a subjective, qualitative activity. 

Cue exploration continued with activity~{\tt (f)}, in which Item~{\tt (PROMPT1.1)} (clarity of the analogy) and Item~{\tt (PROMPT1.3)} (concrete examples) were separately explored through activities~{\tt (a)} and~{\tt (b)}. Cue evaluation using activity~{\tt (c)} and cue selection utilizing activity~{\tt (d)} were performed during activities~{\tt (a)} and~{\tt (b)}. The cues obtained for each of the two items were combined using activity~{\tt (e)}. 

Exploring Item~{\tt (PROMPT1.1)} (clarity of the analogy) conducted the following steps. 

Second, prompting ChatGPT to elaborate on the analogy between the main description of the analogy, such as between topic in poetry and song in dance, created the following related options: {\tt (PROMPT2.1)}~structural foundation as the foundation for the analogy; {\tt (PROMPT2.2)}~emotional expression as both poetry and dance are ways of communicating a certain emotion; {\tt (PROMPT2.3)}~narrative and storytelling; {\tt (PROMPT2.4)}~unity and coherence; and {\tt (PROMPT2.5)}~artistic interpretation as a way of expression for the authors and performers. 

Among the five possibilities, we selected Item~{\tt (PROMPT2.2)}, as improving the emotional expression of the description would also likely increase the emotional impact of the analogy. Hence, the next step combined emotional expression with the idea of body movement in choreography, a detail that was expressed in the paragraph to support the analogy. This combination was expected to improve items~{\tt (PROMPT1.1)} (clarity of analogies) and~{\tt (PROMPT1.2)} (structural flow) due to the more detailed similarity of the emotional outcomes of poetry and dance. 

The third prompt to ChatGPT asked to elaborate on the idea that coordination of body movements in dance is fundamental in conveying emotions to the audience. The produced feedback included the following items: {\tt (PROMPT3.1)}~facial expression; {\tt (PROMPT3.2)}~body language; {\tt (PROMPT3.3)}~timing and rhythms; {\tt (PROMPT3.4)}~spatial dynamics; {\tt (PROMPT3.5)}~partnering and interactions; {\tt (PROMPT3.6)}~narrative and expression; {\tt (PROMPT3.7)}~musical alignment; and {\tt (PROMPT3.8)}~artistic interpretation. The items on body language and spatial dynamics were selected from the list, as it was considered that they connect better than the rest of the items. 

Exploring Item~{\tt (PROMPT1.3)} (concrete examples) originated the next steps.

The fourth prompt to ChatGPT asked for new ideas to further elaborate the above paragraph to explore parallels between the idea of ``Song in Dance'' and the idea ``Topic in Poetry''. The obtained results included the following cues: {\tt (PROMPT4.1)}~historical evolution; {\tt (PROMPT4.2)}~cross-cultural perspectives; {\tt (PROMPT4.3)}~interdisciplinary collaborations; {\tt (PROMPT4.4)}~emotional input; {\tt (PROMPT4.5)} narrative and symbolism; {\tt (PROMPT4.6)}~experimental approaches; {\tt (PROMPT4.7)}~contemporary examples; and {\tt (PROMPT4.8)} educational significance. The following two cues were selected: Cue~{\tt PROMPT4.4} as it helps the convergence with the other parallel exploration thread as emotional expression was also {\tt PROMPT2.2} and {\tt PROMPT4.7}, as the author felt that using contemporary examples not only closely pertains to the explored Item~{\tt PROMPT1.3}, but also is more likely to create an expected emotional outcome from the reader, while being easy to combine with the other emotion-related cues that were selected.  

The fifth prompt asked for an example that illustrates how ``Song in Dance is like Topic in Poetry'' with the answer {\tt PROMPT5.1} referring to the Swan Lake ballet to parallel the song in dance and the topic in poetry. Specifically, the ChatGPT response mentions in music swan princess Odette as a haunting and melancholic character, and black swan Odille as a vibrant and lively persona. In poetry, the response refers to emotional resonance by using the white swan to symbolize purity and innocence and the black swan to represent deception and seduction. Note that the response used a single example, e.g., Swan Lake ballet, to exemplify the analogy. However, this response was considered by the author hard to be incorporated into the paragraph. 

Hence, the sixth prompt asked for a contemporary example for the analogy and the response {\tt PROMPT6.1} indicated the musical Hamilton. Regarding its song, the response mentions the rhythm, wordplay, and tempo of the songs. These are used to describe a broad range of emotions, from excitement to heartache. Regarding poetry, the lyrics describe historical events, characters, and political ideas. The response also indicates the interplay elements between song and poetry, like using fast-paced rap songs and slow ballads to describe specific social situations during the musical action. Similar to the response to the fifth prompt, it was considered that the cues in the response are hard to be incorporated into the paragraph.

The third kind of examples that had to be explored considered previous personal experiences as sources for emotional expression. In addition, the used prompting attempted to combine the desired query with the selected prompts of the other thread, like {\tt PROMPT3.2} (body language) and {\tt PROMPT3.3} (spatial dynamics). Hence, the input to ChatGPT inquired ``how is body movement in dance influenced by previous life experiences?''. The offered results included the following: {\tt PROMPT7.1}~emotional expression; {\tt PROMPT7.2}~cultural background; {\tt PROMPT7.3}~personal narrative; {\tt PROMPT7.4}~physical conditioning; {\tt PROMPT7.5}~psychological factors; {\tt PROMPT7.6} interpersonal relationships; {\tt PROMPT7.7}~life themes; and {\tt PROMPT7.8}~educational background. The author decided to select {\tt PROMPT7.6}, like using time and rhythm to express interpersonal relationships, and {\tt PROMPT7.7}~such as using past personal experiences to inspire a choreography. 

At this point, it became the author's choice to consider the experience of loneliness during learning as the cue to be further pursued during exploring of how to transform the initial paragraph. This ended this case study on using ChatGPT to improve an essay based on an analogy between poetry and dance.

\section {Example 2: Metaphor}

The second experiments considered the next paragraph from another essay by the same author as the essay in the first example~\cite{Sandy2}. The paragraph uses the metaphor of Icarus to present the author's dance experience. 

``Icarus' personality synthesizes my dance experience. I always dashed to the front center of the room to try new styles and tricks in improv situations, choreograph small routines, stumble while attempting a la second turns. I had self-confidence and excitement every time I walked into my regular classes, but I was rarely called out by the teachers, even though I thought I mastered every combination with vivacity. Was I trying too hard or was my confidence too high? Every tear that softened my bed strengthened my desire to perform well, but the same outcome happened every class. Every time I entered a new studio, my liveliness reached higher limits, but nothing new happened. Initially, I assumed that everyone who watched me would be galvanized; my targeted audience was me. Connecting this to the story of Icarus, my hubris was my high self-confidence, which prevented me to spot small imperfections, which probably resulted in the lack of acknowledgement from the teachers. After some time, I started to record and criticize my dance skills. I had to analyze and vanish my small mistakes every class. How can I gesticulate my arms so that I won't have anything to brute about?, I always repeated every time before I recorded myself. It was preparation for masterclasses, when professionals taught us their small routines.''

\subsection {Exploration Trace}

The first prompt to ChatGPT asked to comment on the above paragraph. The obtained response~{\tt PROMPT1.1} assessed the paragraph as being a deeply personal description of the author's journey and struggles in his attempt to become a dancer. It describes the depth and meaning of his personal experience, including his passion, self-awareness, determination, and dedication ({\tt PROMPT1.2}). Moreover, the response indicated his high self-confidence, as a source for motivation but also blinding him for his own imperfections ({\tt PROMPT1.3}). The response also suggested the author's willingness to self-reflect, and to take correction steps to address his shortcomings, suggesting his commitment to growth and mastery ({\tt PROMPT1.4}). Hence, the two main ideas of the paragraph as summarized by ChatGPT's response were the author's high confidence with the risk of making mistakes, and his openness to self-reflection to correct his mistakes. 

Note that the response did not explicitly articulate possible ways of connecting the two ideas, like balancing between self-assurance as an enabler of creative expression and self-reflection for addressing shortcomings, learning, and growth. However, it can be argued that this interpretation of the paragraph is partially incorrect, as Icarus' story focuses on the tragic results of self-confidence but does not discuss using any of its imperfect outcomes to originate improvement and mastery. Hence, {\tt PROMPT1.4} does arguably not have a correspondence in Icarus' story, possibly reducing the soundness of the metaphor. Still, the author felt that Icarus would be a good metaphor for his dance experience, therefore triggering the need for exploration to transform the paragraph to fit better the desired metaphor.              

The pursued exploration decided to independently search for cues starting from three of the concepts mentioned in {\tt PROMPT1.2}, dedication, determination, and passion, as they arguably were the ones having a direct association with the story of Icarus, while self-awareness is the missing part. The idea of {\tt PROMPT1.3}, such as self-confidence as a source for daring motivation but also errors was used to assess the obtained cues, e.g., to decide if the cues support the statement in {\tt PROMPT1.3}. Finally, the idea in {\tt PROMPT1.4} was dropped, as it does not relate well with the mythological story.  

The second prompt asked ChatGPT to suggest specific examples to highlight the author's dedication and unwavering commitment to improving his dance skills. The obtained response included the following ideas: time and effort, like schedule, early morning and late evening practice hours ({\tt PROMPT2.1}); feedback seeking, like accepting active, constructive criticism ({\tt PROMPT2.2}); perseverance through failure, i.e. resilience and push through disappointments, setbacks, and missed opportunities ({\tt PROMPT2.3}); sacrifices made, including the related choices ({\tt PROMPT2.4}); continuous self-improvement, including such examples ({\tt PROMPT2.5}); long-term goals, like dreams and aspiration ({\tt PROMPT2.6}); and passion for the art ({\tt PROMPT2.7}). After the elimination of the feedback that relates to self-improvement and perseverance, which are less connected to Icarus' story, the remaining cues still possible to be explored were as follows: {\tt (PROMPT2.1)}, {\tt (PROMPT2.4)}, {\tt (PROMPT2.6)}, and {\tt (PROMPT2.7)}. We selected the cue of a long-term goal ({\tt PROMPT2.6}) among the four remaining. While Icarus did not explicitly state a long-term goal in his story, still, it is still possible to identify some goal for his quest to fly as close as possible to the sun, and hence these cues still maintain the soundness of the metaphor.

The third prompt asked ChatGPT to highlight determination, as it was part of the selected {\tt PROMPT1.2}. The response offered the next eight cues: consistent effort, like trying out new things and overcoming obstacles ({\tt PROMPT3.1}); relentless self-reflection ({\tt PROMPT3.2}); perseverance among frustration, like self-doubt and emotions to fuel harder work  ({\tt PROMPT3.3}); trial and error ({\tt PROMPT3.4}); continuous learning ({\tt PROMPT3.5}); unaware self-belief, including never losing faith ({\tt PROMPT3.6}); goal-oriented mindset ({\tt PROMPT3.7}); and long-term commitment ({\tt PROMPT3.8}). Similarly, to the responses for the second prompt, after eliminating cues about self-reflection and continuous learning, the cues in ({\tt PROMPT3.7}) and ({\tt PROMPT3.8}) were very similar to the previously selected cue ({\tt PROMPT2.6}). Moreover, cue ({\tt PROMPT3.1}) mentioned the desire to try out new things, which connects well to the story of Icarus, and is also explicitly mentioned in the original paragraph. Thus, these three cues were selected for further exploration. However, cues ({\tt PROMPT3.3}), ({\tt PROMPT3.4}), and ({\tt PROMPT3.6}) could have been also considered, but while they are sound within the Icarus' story, they are less linked to the initial paragraph, and thus, the needed paragraph transformations would be more significant.  

\begin{figure*}
\centering
\includegraphics[width=6.0in]{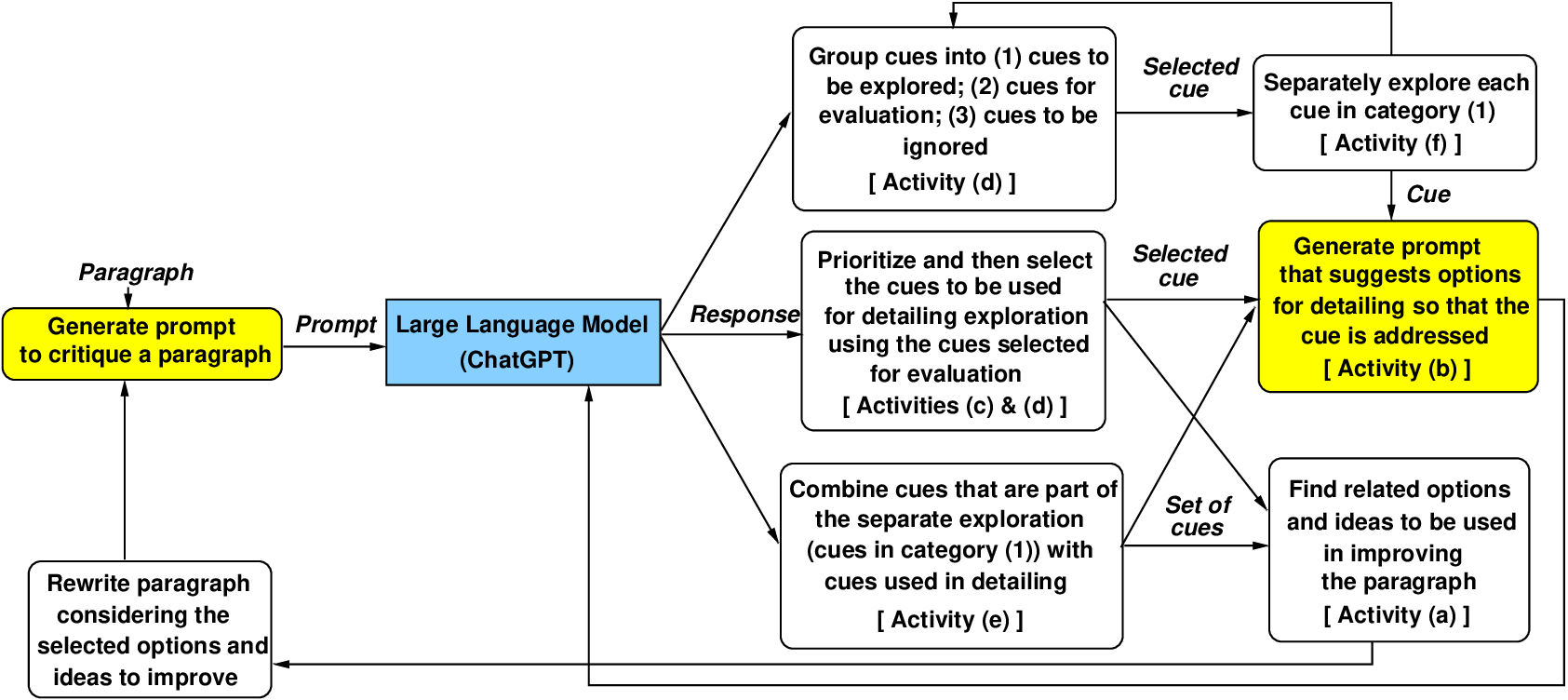}
\caption{Activities and flow in the algorithmic approach for paragraph improvement}
\label{prompting1}
\end{figure*}

The fourth prompt requested ChatGPT to detail passion. Passion was part of the cues in {\tt PROMPT1.2}. The response indicated that passion can be highlighted in the paragraph by emphasizing the author's deep emotional connection to dance and his unwavering love for the art form. The following eight specific ideas were offered: expressive language ({\tt PROMPT4.1}); physical sensations ({\tt PROMPT4.2}); sensory details, like the sound of the music filling a room ({\tt PROMPT4.3}); moments of euphoria ({\tt PROMPT4.4}); personal sacrifice ({\tt PROMPT4.5}); artistic inspiration, i.e. how other's work inspires passion ({\tt PROMPT4.6}); intrinsic motivation, e.g., pure joy ({\tt PROMPT4.7}); and dreams and applications ({\tt PROMPT4.8}).  

This ended this case study on using ChatGPT to improve an essay using a well-known metaphor to present a personal experience.

\section {Algorithmic Approach}

Figure~\ref{prompting1} depicts the algorithmic approach used in sections III and IV to guide the restructuring of a paragraph to better address the intention of the author, like improving the identity component of the writing. The figure includes activities (a)-(f) in Section~III shown as white boxes, the prompt generation steps as yellow boxes, and the LLM (e.g., ChatGPT) as a blue box. The approach is losely based on the ideas of the cognitive architecture discussed in~\cite{Li2018}. The algorithmic approach is as follows.

The first step uses a prompt to ask for a critique of an initial paragraph that must be improved. The utilized prompts followed the templates {\tt ``Comment the paragraph ...''}, {\tt ``Critique the paragraph ...''}, and {\tt ``Expand the paragraph ...''}. The responses produced by the LLM, e.g., ChatGPT, can include the following elements: (i)~paragraphs that comment on the strength and limitations of the input paragraph, including main nouns in the paragraph, i.e. nouns used to express the main ideas, and attributes and adverbs, like ``effectively conveys'', ``Beautifully captures'', ``deeply personal''; (ii)~enumeration of specific points on how to improve the paragraph, and short description and examples for each item; (iii)~rewording through summaries of the paragraph; and (iv)~separation of the distinct ideas that are discussed in the paragraph. The responses produce cues that become the starting points for the following exploration process. 

Depending on the current state of the exploration process, the cues in the response can go through three different processing steps, as shown in Figure~\ref{prompting1}. 

(1) Initially, the cues in the response are grouped into three categories: (i) cues to be further explored, (ii) cues for evaluation, and (iii) cues perceived to be less important, and will be ignored. This is Activity~(d) in the activity list enumerated in Section~III. As mentioned in Section~III.A, cues {\tt PROMPT1.1} and {\tt PROMPT1.3} were selected for further exploration, e.g., they formed category~(i), and cues {\tt PROMPT1.2} and {\tt PROMPT1.4} were used for evaluation, i.e. they formed category~(ii). This is a personal choice of the author depending on how he / she intends to frame and expand the paragraph. The cues selected to be further explored are separately considered, one-by-one by the parallel exploration loop of the approach.    

(2) Alternatively, for each of the cues separately searched by the parallel exploration loop, the LLM responses are processed by prioritizing the selected cues. They implement activities~(c) and~(d), and serve to detail the exploration process. The used prompts asked for elaborations, examples, including contemporary examples, and summaries. The obtained results usually include three parts:  (i)~a broad statement about the idea of the response, (ii)~a list of possibilities, and (iii)~a summary that includes a list of the main concepts of the response, which can then be part of the set of possible cues. The list of possibilities can be at different levels of details depending on the prompt. For example, the list can refer to the parameters that define it, examples of choices, like rush of adrenaline or joy of movement to convey passion, desired goals or attributes to be achieved, e.g., achieving a calm demeanor on stage, and concrete examples, i.e. sudden collapse of a dancer to the floor to show despair. Examples of the related cues are {\tt ``How do you highlight Selected\_Cue in this paragraph''}, {\tt ``What are famous individuals that showed Selected\_Cue?''}, or {\tt ``What are famous literary characters that exemplify Selected\_Cue''}. {\tt Selected\_Cue} is the cue selected to guide detailing. 

(3) Finally, the cues that are separately used to guide an iteration of the parallel exploration loop are combined with cues that are utilized in detailing. This step corresponds to Activity~(e). The structures of the utilized prompts included the following cases: {\tt ``How is Boader\_Cue influenced by Detailed\_Cue?''}, e.g. ``How is body movement influenced by previous life experiences?'', {\tt ``How does Detailed\_Cue convey Goal\_Cue''}, i.e. ``How does body movement convey emotions?'', {\tt ``How do you express Detailed\_Cue in Broader\_Cue?''}, like ``How do you express social distance in dance'', or {\tt ``How do you balance Cue and Cue?''}, such as ``How do you balance self-determination and self-improvement?'', where both cues can be at the same or at different levels of detailing.

The cues produced through Processing Step~(1) are separately explored one at a time to realize the outer exploration loop. It is activity~(f) in Figure~\ref{prompting1}. They offer a breadth-oriented search along the main cues, so that each of the cues receives the highest priority during exploration. For each of these cues, detailing adds specificity by using prompts that pertain to category~(2). 

The convergence of the parallel and detailed exploration steps is achieved by using cues that are in category~(3), as described before.   

The author decides when to rewrite the paragraph based on the resulting cues, which is then continued by re-starting the exploration loop by generating a prompt to critique the paragraph. Exploration continues until it considers that the paragraph description is sufficient.

\section {Evaluation}

\begin{figure*}
\centering
\includegraphics[width=5.4in]{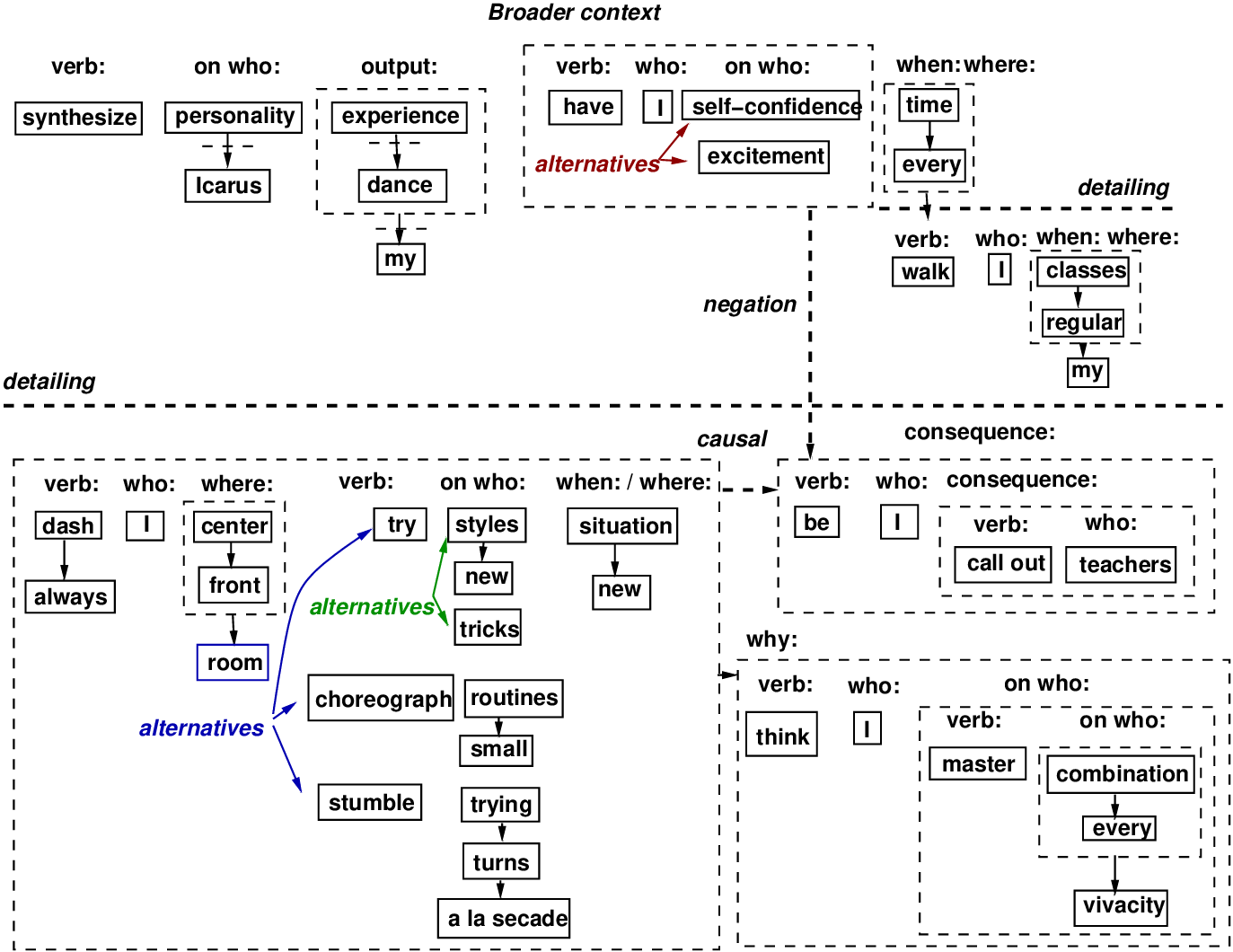}
\caption{Activities and flow in the algorithmic approach for paragraph improvement}
\label{prompting2}
\end{figure*}

Figure~\ref{prompting2} illustrates the representation considered for the authored paragraphs, prompts, and the LLM-generated responses. More details about the representation are offered in~\cite{Doboli2023a, Doboli2023c}. The representation comprises of clusters of related ideas (e.g., sentences), so that the clusters describe the context of a sentence or paragraph, as well as its detailing, causal, supporting, or negation relationships between the clusters. Note that relationships can be explicitly expressed in the paragraph or are part of the process of understanding a paragraph. A simple sentence understanding procedure has been proposed in~\cite{Doboli2023b}, but future work must significantly expand this approach.   

Each sentence of a text description is parsed into a representation that indicates the roles of the words in the sentence~\cite{Doboli2021, Doboli2023b, Fauconnier2002}: as shown in Figure~\ref{prompting2}, words are placed into distinct ``bins'' depending on their roles in a sentence: verb, what, who, where, when, on~who (is the action of the verb exerted), outputs (of the action), for who, and why roles. These are the current roles that we identified, but other roles are possible too. Attributes and adverbs can qualify nouns and verbs, like attribute ``regular'' being associated to noun ``classes'' in the figure.

In addition, the representation in Figure~\ref{prompting2} indicates the relationships that exist between the sentences of a paragraph. First, {\em Broader context} describes the sentences that describe the environment, e.g., goals, assumptions, under which the following sentences should be processed. The left sentence in the figure states the most general summary of the author's intention, which is to use the story of Icarus to describe his own experience in dance. The right sentence of the broader context details the specifics of how the metaphor was selected, like to express the author's self-confidence. {\em Detailing} is a second relationship between sentences. Third, {\em alternative} relationships present different options, like the sentences ``choreograph small routines'' and ``stumble trying turns a la secade''. These sentences serve the same purpose in the overall paragraph. Fourth, {\em causal} relationships indicate situations when a sentence is a consequence of a previous sentence or group of sentences, like the earlier sentences determine the consequence. Fifth, {\em negation} relationships sentences that contradict another sentence. Sixth, {\em connected} relationships indicate that sentences are related to each other due to their similarities being above a certain threshold.

A set of metrics were defined using the proposed representation for input paragraphs, prompts, and LLM responses. The metrics refer to the following aspects:
\begin{enumerate}
\item
{\em Explored space}: These metrics refer to the breadth and depth of the explored ideas, e.g., LLM responses, related to an author's paragraph. It refers to the degrees in which breadth and depths are expanded starting from the initial representation, including the distributions of the path lengths of connections between the LLM responses and paragraphs~\cite{Liu2021, Liu2023}. Metrics also express the centrality of ideas, such as the degree to which they are related to the other ideas~\cite{Liu2021, Liu2023}. It also presents ideas, nouns, verbs that are not being forgotten during exploration.

\item
{\em Unexplored cases}: These metrics describe ideas, nouns, verbs that are forgotten during exploration.

\item
{\em Degree of connection}: These metrics describe the amount, nature, soundness, and relevance of the relationships between paragraphs, prompts, and LLM responses.  

\item
{\em Unconnected ideas}: The metrics refer to ideas which are less connected in general or less connected through a certain kind of relationships. 

\item
{\em Idea inconsistencies}: The metrics express connections between ideas, so that the resulting structure is unsound. 

\item
{\em Idea flow}: The metrics describe the paths of the relationships between ideas~\cite{Liu2023}. These metrics suggest the likelihood of ideas to be correctly understood. For example, ideas linked through long sequences of causal relationships are likely more difficult to understand than well-connected ideas, as unambiguously figuring out the connections between ideas in the same way might be hard as different users might produce distinct interpretations.

\end{enumerate}

\begin{figure*}
\centering
\includegraphics[width=6.4in]{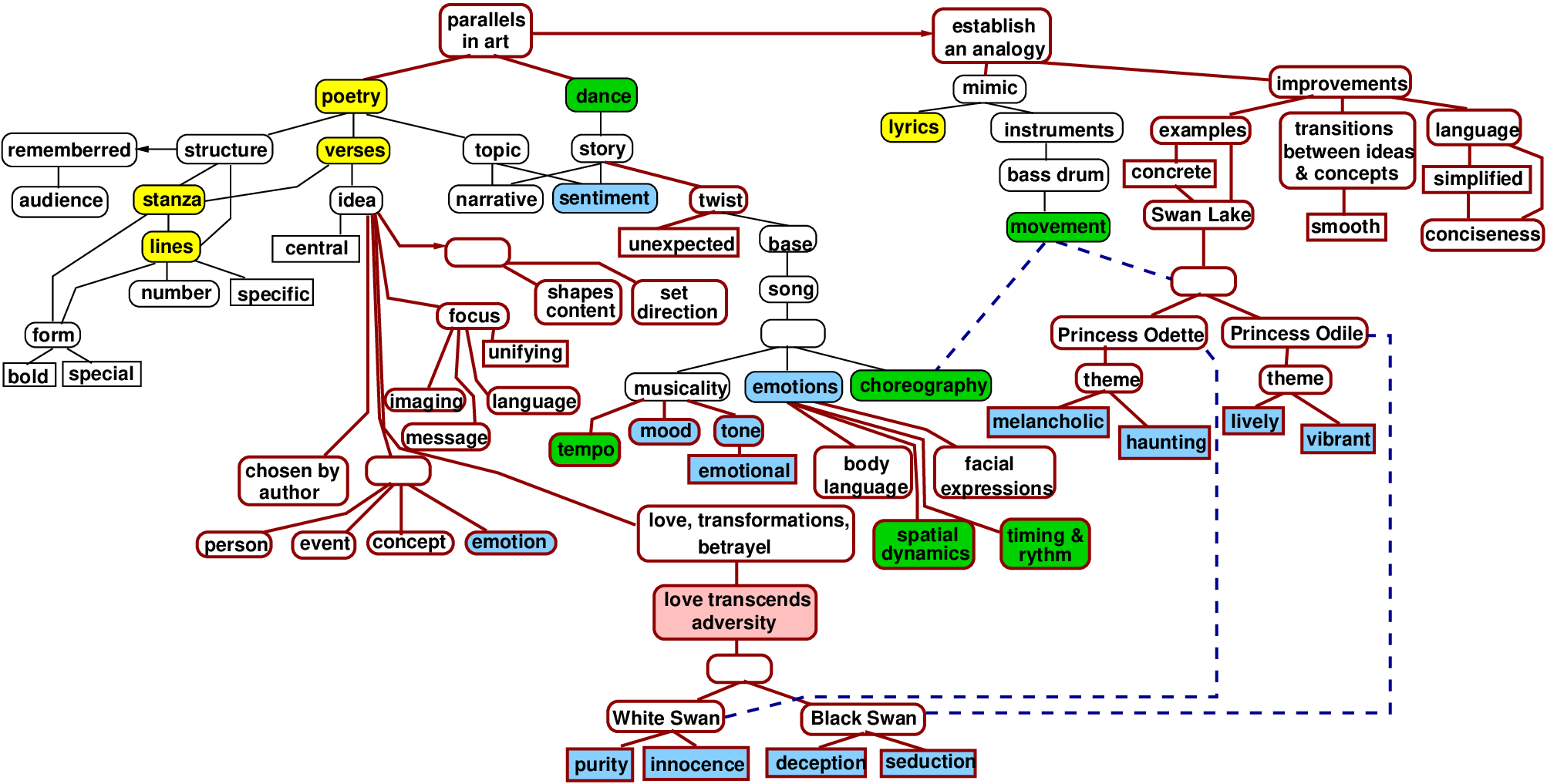}
\caption{Concept Graph for the analogy discussed in Section~III}
\label{prompting3}
\end{figure*}

\section {Experiments}

\subsection {Analogy in Section~III}

Figure~\ref{prompting3} depicts the Concept Graph (CG) for the case study discussed in Section~III on the story discussing the analogy between poetry and dancing. The graph was built using the model discussed in~\cite{Doboli2015, Ferent2013b}. The figure shows the utilized concepts, and their connections (through undirected arcs) depending on their level of abstraction. For example, concept ``poetry'' includes three related concepts, hence at a lower level of abstraction as they express facets of poetry, ``structure'', ``verses'', and ``topic''. The directed edges express  a goal, such as concept ``structure'' is something that the audience should remember. The concepts with black margins are those that occur in the original write-up. The concepts with red margins are those added by the LLM in response to the offered prompts. The edges shown with dashed line are relationships between concepts, but which were not explicitly expressed in the write-up. 

As shown in Figure~\ref{prompting3}, the initial write-up included twenty-four concepts linked in twelve paths. The lengths of the paths are as follows: one path has length two, five paths have length three, three paths have length four, and four paths have length five. Hence, the corresponding CG is wider and less deep. The responses of the LLM changed the CCG as follows: thirty new concepts were included, nineteen new paths were introduced, and the additional path length are as follows: two paths have a length of four, five paths have a length of five, two paths have a length of six, four paths have a length of seven, and six paths have a length of eight. Hence, the shape of the CG changed into deeper structure.  

Figure~\ref{prompting3} indicates that the used concepts can be grouped based on their relatedness into three clusters shown with different colors: a cluster is formed by concepts ``poetry'', ``verses'', ``stanza'', ``lyrics'', and ``lines'' (color yellow);
a cluster includes concepts ``dance'', ``movement'', ``choreography'', ``tempo'', ``spatial dynamics'', and ``timing \& rhythm'' (color green), and a cluster has concepts ``sentiment'', ``mood'', ``emotional tone'', ``emotions'', ``melancholic theme'', ``haunting theme'', ``lively theme'', ``vibrant theme'', ``purity'', ``innocence'', ``deception'', and ``seduction'' (color blue). The last cluster is the largest, which suggests that discussing emotions is a main topic of the write-up.   

The figure shows the unexplored cases, like the idea ``love transcends adversity'' and the concept ``unexplored twist''. Note that the first idea represents a possible example of a central idea for the write-up to support the description of analogies between poetry and music. However, the idea was not mentioned by the author in the initial write-up. Other unexplored concepts, even though arguably less central to the expressed ideas, are the ideas of ``smooth transitions between ideas and concepts'' and ``conciseness'' through ``simplified language''. 

The most connected concepts is ``central idea'', which is directly linked to thirteen concepts, like concepts ``focus'', ``chosen by author'', and so on. Other concepts more connected than others are ``song'', ``musicality'', ``emotions'', and ``choreography'' related to dance, and concepts ``poetry'', ``structure'', ``verses'', and ``topic'' for poetry. Thus, based on their connectivity, the analysis suggests that the main concepts refer to movement, emotions and their parallels with poetry. 

Some of the unconnected ideas are shown with blue dashed lines in Figure~\ref{prompting3}. They include concepts ``movement'' and ``choreography'', and concepts ``central idea'' and ``narrative''. Note that these concepts occur in different parts of the CG. 

The shown Concept Graph also includes inconsistencies, like between the features mentioned for concept ``Princess Odette'', like haunting and melancholic, and for concept ``Princess Odile'', such as vibrant and lively. However, their analogous concepts have other features, such as purity and innocence for concept ``White Swan'', and deception and seduction for concept ``Black Swan''. There is some inconsistency between attributes that should match, like deception and seduction (suggesting a negative sentiment) and vibrant and lively (indicating a positive sentiment). These inconsistencies can weaken the soundness of the described analogy and create confusion.    

\begin{figure*}
\centering
\includegraphics[width=5.6in]{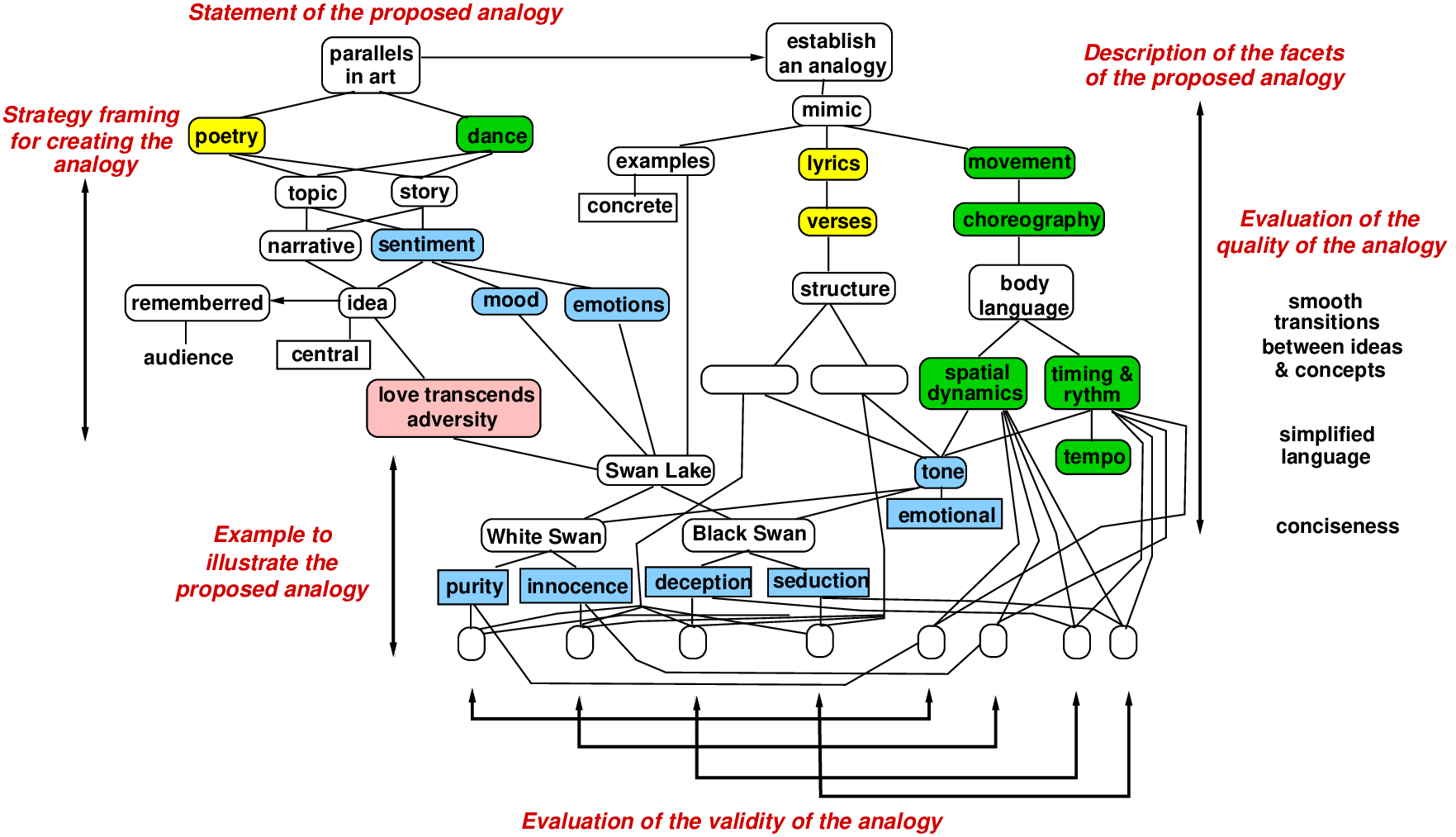}
\caption{Concept Graph (CG) for the modified write-up to improve the analogy construction while presenting a central idea}
\label{prompting4}
\end{figure*}

Figure~\ref{prompting4} depicts the CG for the modified paragraph to improve the description of the analogy between poetry and dance, while adding a central idea that reflects the author's identity. Only concepts present in Figure~\ref{prompting3} were used for creating the modification. The selected central idea is ``love transcends adversity'' and is highlighted in pink in the figure. The modified CG includes five activities shown in red in the figure. The main parts of the CG corresponding to the five activities are as follows.

The statement of the proposed analogy, as shown at the top of the figure, is to establish an analogy between two artistic domains, such as poetry and dance. The top, left sub-graph connects the two main features of poetry and dance, like topic and story, and the two instantiated concepts, e.g., narrative and sentiment, to the central idea, on which the analogy description is built. Hence, the sub-graph frames the main strategy for building the analogy. This part corresponds to the beginning of the new write-up.  

The top, right sub-graph describes the utilized facets to construct the analogy. Establishing the analogy requires mimicking the main facets in poetry and dancing. One specific way of mimicking is to utilize concrete examples. The two next detailing for concept ``lyrics'', i.e. concepts ``verses'' and ``structure'' should match the two corresponding detailing of concept ``movement'', e.g., concepts ``choreography'' and ``body language''. However, the next detailing of the later concept, like concepts ``spatial dynamics'', ``timing \& rhythm'', and ``tempo'' do not have matching detailing concepts on the side of concept ``lyrics''. These missing concepts were shown as white ovals in the figure and suggest additional concepts that must be included to the new write-up to improve the analogy making. 

The bottom, middle sub-graph presents the examples used to illustrate the proposed analogy. The selected central idea, e.g., ``love transcends adversity'' lead to picking the two main characters of Swan Lake ballet to present the analogy between poetry and dancing. These characters are White Swan and Black Swan, each described by two attributes, like purity and innocence, and deception and seduction, respectively. This leads to the four descriptions of the four attributes for poetry, and similarly, the description of the same four attributes for dancing. The validity of the analogy is evaluated based on the degree to which the four pairs of corresponding attributes describe a good matching. 

Finally, the middle, right part of the figure shows the evaluation of the quality of the constructed analogy, which includes three aspects: smooth transitions between ideas and concepts, simplified language, and conciseness.  

\begin{figure*}
\centering
\includegraphics[width=6.3in]{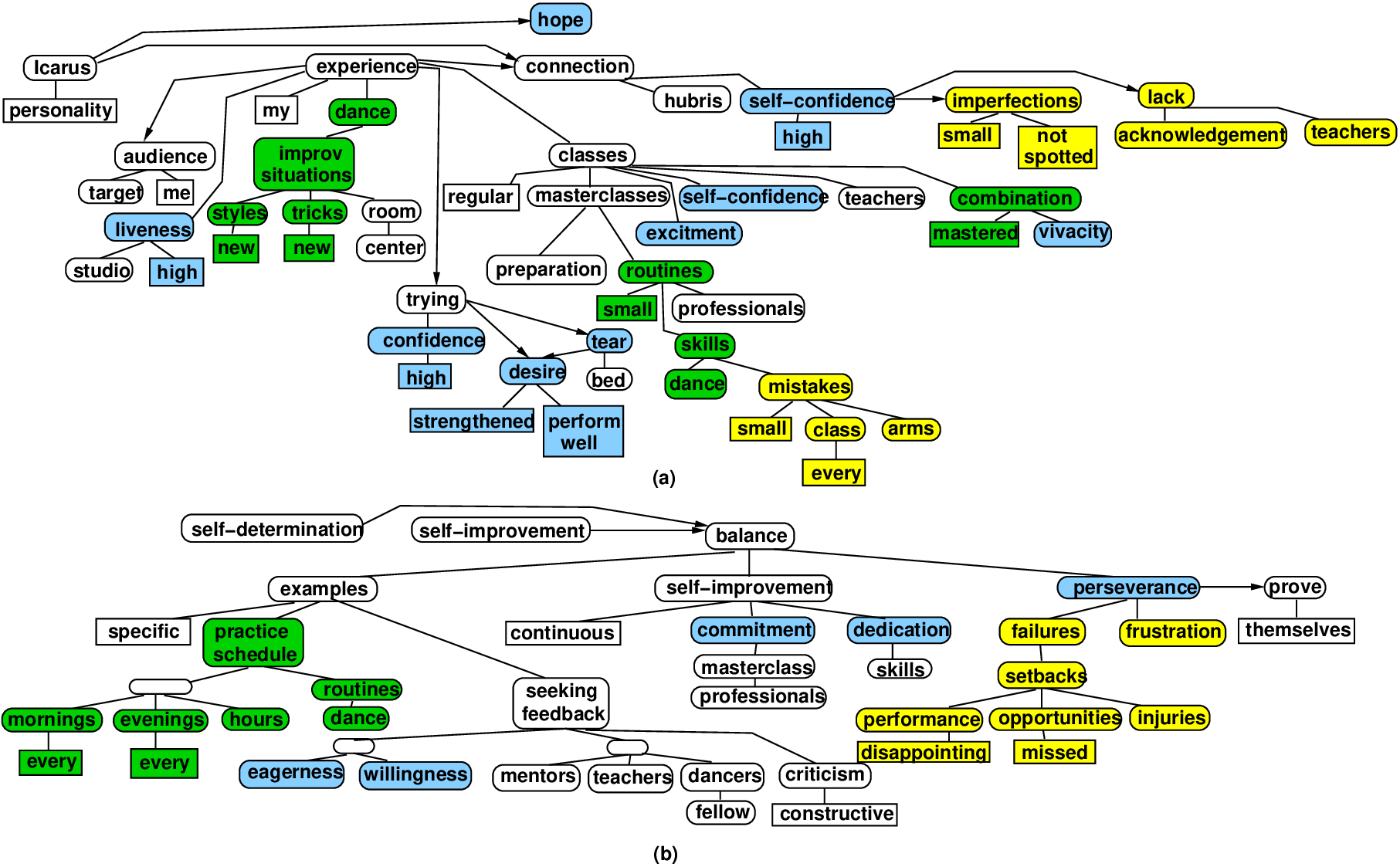}
\caption{Concept Graph (CG) for the metaphor discussed in Section~IV: (a) CG of the author's write-up and (b) CG fragment of the LLM responses}
\label{prompting5}
\end{figure*}

\subsection {Metaphor in Section~IV}

Figure~\ref{prompting5} depicts the CG of the metaphor-based example discussed in Section~IV. Figure~\ref{prompting5}(a) is the CG of the initial write-up by the author, and Figure~\ref{prompting5}(b) represents the CG for the LLM for some of the responses for given prompts. The part not shown is similar.   

The graph in Figure~\ref{prompting5}(a) includes thirty nine concepts linked in twenty three paths, and the graph in Figure~\ref{prompting5}(b) adds more than thirty two concepts in twenty paths (the figure shows only a fragment of the corresponding CG). The path lengths of the graph in Figure~\ref{prompting5}(a) are as follows: two path are of length one, six paths have length two, eight paths have length three, four paths are of length four, one path has length five, and two path are of length six. The path length distribution for the graph fragment in Figure~\ref{prompting5}(b) is as follows: two paths are of length one, one path has length two, nine paths are of length three, and eight paths have length four. The path lengths suggest that the two graphs are shallower, hence are more breadth oriented, than the graph in Figure~\ref{prompting3}. 

The shallower structures in Figure~\ref{prompting5} suggest that the description is less detailed as compared to the analogy-based paragraph in Section~III. The concepts were grouped into four clusters: concepts related to the author's experience, like ``dance'', ``trying with confidence'', ``improvisations'', ``new styles'', ``new tricks'', and ``mastered combinations'' (shown with green color), concepts about the author's high self-confidence, e.g., ``vivacity'', ``excitement'', ``strengthened desire'', ``desire to perform well'', ``hope'', and ``high liveness (depicted with blue color), concepts referring to the experienced challenges, i.e. concepts ``imperfections'', ``lack of acknowledgement'', ``small mistakes'', ``arms'', and ``every class'' (shown with yellow color), and the rest of concepts, which serve a supporting role in the description (white color).  The used concepts ``new styles'' and ``new tricks'' suggest the intention to go beyond the traditional norms. Note that the first three clusters include about the same number of concepts, which suggests that the same emphasis was given to the three related ideas. 

There is no direct statement of a central idea, other than that saying that the author's experience parallels the story of Icarus. Moreover, concept ``Icarus'' in Figure~\ref{prompting5} is unconnected to the rest of the graph, even though it is expected to play an important role in the narrative. Other unconnected concepts are ``imperfections'', ``failures'', and ``mistakes'', which similarly are important in the story of Icarus. Another less connected concepts is ``self-confidence'', such as the concept is less detailed in the graph. Still, the repetition of concepts ``high self-confidence'' and ``imperfections'' indirectly suggests that the relationship between the two concepts is an important message. The existence of these relationships is also supported by having concepts of colors green, blue, and yellow connected as part of the same path. Note that the graph in Figure~\ref{prompting5}(b) of the LLM feedback indicates the existence of a central idea, which is the balance between self-determination and self-improvement, even though arguably a more precise statement of the central idea would be the balance between self-determination and experienced setbacks, with self-improvement being the consequence of the opposite tension between the two. 

Hence, the reader must infer the using of the metaphor of Icarus to understand the intention of the author under the paragraph's statements which follow a structure that refers to a dancing activity, the author's excitement and self-confidence, and the encountered challenges. There is no explicit suggestion on how to interpret the suggested metaphor to understand the author's story, e.g., to solve the ``puzzle'' produced by the metaphor. This is likely to produce different interpretations of the paragraph.     

\section {Conclusions}

This paper presents a novel computational approach to improve write-ups using the responses offered by Large Language Models (LLMs) in response to a sequence of prompts. The work focuses on write-ups that reflect an author's identity, like personal experiences. The sought improvements refer to the clear framing of the central ideas, the credibility of the presented ideas, e.g., the correctness of any suggested analogies and metaphors, referring to examples in the presentation, or the idea flow of the narrative. Six activities were identified based on two write-ups, one presenting an analogy and one using a metaphor, to support the write-up improvement process. The activities refer to finding related concepts and ideas, detailing, LLM feedback (cue) evaluation, cue selection for further exploration, combining cues and write-up ideas, and separate exploration of different ideas. The computational approach uses the six activities in an iterative exploration loop to generate and select cues, and then combine them with the considered write-up. A set of metrics to evaluate the modified write-ups is also discussed in the paper.

The experiments focus on using the computational approach to improve two separate identity-oriented write-ups, one based on the analogy between poetry and dancing, and the second using the story of Icarus in Greek mythology to discuss the balance between the author's self-confidence and failures in dancing. Both situations represent instances of puzzle-solving cases, in which matching relationships, i.e. isomorphisms, must be established between the ideas, nouns, and verbs of the write-ups and the supporting analogy or metaphor. Concept Graphs (CGs) were built for each case to reflect the connections between the concepts of the write-ups and the feedback produced by the LLM. The CG for the analogy-based write-up is deeper, thus includes more detailing, as compared to the metaphor-based write-up, which is more shallow, hence mainly reflects possible alternatives. CGs also indicate opportunities to improve a write-up by highlighting situations in which the central idea is not well connected to the rest of the write-up, concepts that are not sufficiently detailed, or write-up structures that are not aiding the puzzle solving process needed to understand the write-up.

\bibliographystyle{ACM-Reference-Format}
\bibliography{wp}

\end{document}